\documentclass{egpubl}
 
\JournalSubmission    
\usepackage[T1]{fontenc}
\usepackage{dfadobe}  
\usepackage{xspace}
\usepackage{xcolor}
\usepackage{amsmath,amssymb} 
\usepackage{booktabs} 
\usepackage{mathtools} 

\usepackage{cite}  
\BibtexOrBiblatex
\electronicVersion
\PrintedOrElectronic
\ifpdf \usepackage[pdftex]{graphicx} \pdfcompresslevel=9
\else \usepackage[dvips]{graphicx} \fi

\usepackage{egweblnk} 

\newcommand{\data}[1]{\textsc{#1}\xspace}
\newcommand{\ABC}{\data{ABC}}
\newcommand{\famous}{\data{Famous}}
\newcommand{\real}{\data{Real}}
\newcommand{\thingi}{\data{Thingi10k}}
\newcommand{\name}{\mbox{\textsc{PPSurf}}\xspace}

\newcommand{\diff}[1]{#1}  

\title[\name: Combining Patches and Point Convolutions for Detailed Surface Reconstruction]%
      {\name: Combining Patches and Point Convolutions \\ for Detailed Surface Reconstruction}

\author[P. Erler \& L. Fuentes \& P. Hermosilla \& P. Guerrero \& R. Pajarola \& M. Wimmer]
{\parbox{\textwidth}{\centering 
P. Erler$^{1}$\orcid{0000-0002-2790-9279}
and L. Fuentes-Perez$^{2}$\orcid{0000-0003-1096-2871} 
and P. Hermosilla$^{1}$\orcid{0000-0003-3586-4741}
and P. Guerrero$^{3}$\orcid{0000-0002-7568-2849}
and R. Pajarola$^{2}$\orcid{0000-0002-6724-526X}
and M. Wimmer$^{1}$\orcid{0000-0002-9370-2663}
        }
        \\
{\parbox{\textwidth}{\centering 
$^1$TU Wien, Austria\\
$^2$University of Z\"urich, Switzerland\\
$^3$Adobe Research, United Kingdom
       }
}
}

\begin{document}

\teaser{
    \includegraphics[width=\linewidth]{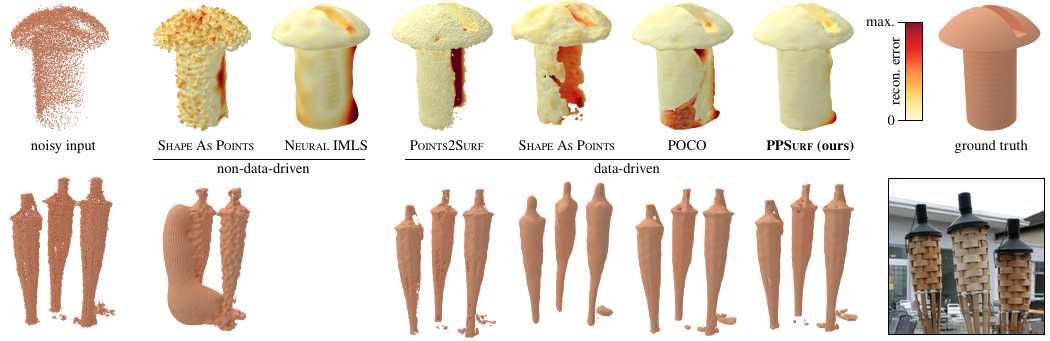}
    \centering
    \caption{
    We present \name, a method to reconstruct surfaces from noisy point clouds. Unlike previous methods, our approach combines two strong data-driven priors, one prior over local surface details, and a second prior over the coarse shape of larger surface regions. This makes \name robust to noise, while reconstructing surface detail better than current methods.
    }
    \label{fig:teaser}
}

\maketitle

\begin{abstract}
3D surface reconstruction from point clouds is a key step in areas such as content creation, archaeology, digital cultural heritage, and engineering. Current approaches either try to optimize a non-data-driven surface representation to fit the points, or learn a data-driven prior over the distribution of commonly occurring surfaces and how they correlate with potentially noisy point
clouds. Data-driven methods enable robust handling of noise and typically either focus on a \emph{global} or a \emph{local} prior, which trade-off between robustness to noise on the global end and surface detail preservation on the local end. We propose \name as a method that combines a global prior based on point convolutions and a local prior based on processing local point cloud patches. We show that this approach is robust to noise while recovering surface details more accurately than the current state-of-the-art.
%
Our source code, pre-trained model and dataset
\diff{are available at: \url{https://github.com/cg-tuwien/ppsurf}}


\end{abstract}

\footnotetext{\tiny Published in Computer Graphics Forum (Jan 2024): \\  \url{https://onlinelibrary.wiley.com/doi/10.1111/cgf.15000}}

\section{Introduction}

3D surface reconstruction from point clouds is a key step for workflows in areas such as content creation, archaeology, digital cultural heritage, and engineering, to convert raw 3D point scan data, like casual RGBD (color and depth) mobile phone images or more accurate range scans (e.g., from laser range scanner), to surface-based 3D object representations that can be used in downstream applications.

Given the large practical interest, surface reconstruction has become a central problem in computer graphics and vision research. The problem is generally ill-defined, as different surfaces may correspond to similar point clouds. However, several approaches have been proposed to tackle this ambiguity. One research direction attempts to optimize surface representations with strong \emph{non-data-driven} inductive biases to fit the point cloud~\cite{kazhdan2006poisson, williams2019deep, Peng2021SAP, NeuralPull2021baorui}. This resolves the ambiguity, but is susceptible to deteriorating conditions of the input points, such as scan noise or regions with missing points, which cannot easily be corrected using a fixed inductive bias.
Another line of research focuses on learning \emph{data-driven} priors, usually over the distribution of commonly occurring surfaces and how they correlate with potentially noisy point clouds~\cite{park2019deepsdf, Points2Surf2020, Peng2021SAP, Boulch_2022_CVPR}. The surface reconstruction ambiguity can then be resolved by finding a surface that has a high probability for the given point cloud under the learned prior. The prior in these data-driven methods can range from \emph{global}, where the prior captures a distribution over full 3D object surfaces, to \emph{local}, where the prior captures the distribution over local surface patches. Global priors are the least susceptible to noise and missing points, but have limited capability to capture fine local details.
Local priors, on the other hand, can capture such fine details accurately, but are more susceptible to strong noise and missing points. Existing methods mostly focus their prior on a small range in this global-local spectrum. For example, DeepSDF~\cite{park2019deepsdf} uses a global prior, Points2Surf~\cite{Points2Surf2020} mostly focuses on a local prior, while POCO's point convolutions~\cite{Boulch_2022_CVPR} learn a prior in the medium range that is reasonably robust to deteriorating conditions, but still struggles to accurately capture local detail.

We propose \name as a method that covers a wider range in the global-local spectrum of priors, by combining the local prior of a patch-based method like Points2Surf with a more global prior of a point convolution-based method like POCO. For this purpose, we design an architecture that has two branches: the first branch is based on POCO~\cite{Boulch_2022_CVPR} and provides a global prior by applying several layers of point convolutions to a sparse set of support points. To reconstruct geometric details more accurately, we merge features from this first branch with features from a second branch, which processes a local patch of points with PointNet~\cite{qi2016pointnet}. We additionally discovered that modifying the architecture of PointNet by replacing the sum aggregation with an attention-based aggregation improves performance. This results in a method that is robust to noise and missing points, while preserving details more accurately than previous methods.


In our experiments, we compare \name to several previous state-of-the-art methods, both data-driven and non-data-driven, on synthetic as well as real-world data, and demonstrate improved performance on both in-distribution, and out-of-distribution surface reconstruction tasks.











\section{Related Work}
 
Surface reconstruction from point clouds is an active area of research.
We distinguish between \emph{data-driven} methods that train on a large dataset, and \emph{non-data-driven} methods that do not use machine learning or overfit to a single shape.

\paragraph*{Non-data-driven methods.} Poisson reconstruction~\cite{kazhdan2006poisson,kazhdan2013screened} has for many years been the gold standard of non-data-driven approaches.
Recent works have suggested optimizing the parameters of a neural network to predict the signed distance to the surface~\cite{sal2020atzmon, sitzmann2019siren, atzmon2021sald} directly from a \emph{single} point cloud.
In particular, Atzmon and Lipman~\cite{sal2020atzmon} introduced this concept for unoriented point clouds.
They optimized the parameters of the neural network with a sign-agnostic loss and a geometric initialization of its parameters.
Gropp et al.~\cite{gropp2020igr} and Atzmon and Lipman~\cite{atzmon2021sald} followed up on this work and included a gradient regularization in the loss.
Later, Ma et al.~\cite{NeuralPull2021baorui} introduced Neural-Pull, an optimization objective that uses directly the gradient of the optimized SDF to move the query points to the closest point in the input point cloud.
In follow-up work, this approach was extended 
by incorporating a network to classify a point being on the surface or not~~\cite{NeuralTPS2023}, and an additional loss that aligns the gradient direction between different level sets of the SDF~~\cite{Baorui2023Towards}.
In order to improve the quality of the final SDF, Yifan et al.~\cite{yifan2020isopoints} and Zhou et al.~\cite{Zhou2022CAP-UDF} proposed to iteratively increase the input point cloud with points sampled from the optimized SDF in the previous iteration.
A different approach was proposed by Peng et al.~\cite{Peng2021SAP} (also used in LION~\cite{zeng2022lion}),
based on a differentiable Poisson Surface Reconstruction operation that could be used for optimization-based or learned reconstructions.
Differently from previous methods, the set of points in the surface is optimized through the differentiable reconstruction instead of a neural network representing the SDF.
Lin et al. proposed a parametric Gauss formula for reconstruction~\cite{lin2023gaussformula}, which has quadratic complexity in memory leading to prohibitive costs for larger point clouds.
VIPSS by Huang et al.~\cite{huang2019vipss} formulates reconstruction as a constrained quadratic optimization problem.
iPSR by Hou et al.~\cite{hou2022ipsr} uses an iterative approach to Poisson reconstruction that improves the surface more and more, while removing the need to be given point normals.
IsoPoisson by Xiao et al.~\cite{xiao2023isopoisson} incorporate an isovalue constraint to the Poisson equation, which helps with consistent normal orientation and consequently improved reconstruction.

Non-data-driven methods are sensitive to noise, which is usually present in real 3D scans.
In order to address this limitation to some extent, a recent pre-print from Wang et al.~\cite{wang2022NeuralIMLS} proposed Neural-IMLS, a non-data-driven method that regularizes the smoothness of surface normals using an MLP with limited capacity. While this produces smooth surfaces, it also loses some geometric detail due to this non-data-driven regularization.
Noise to Noise Mapping by Baorui et al.~\cite{Baorui2023Noise2NoiseMapping} focuses on the reconstruction of noisy point clouds in an unsupervised overfitting scheme.
Additionally, these methods require significant reconstruction times due to the optimization being performed for each shape individually, which can be a limiting factor for large scans.

\paragraph*{Data-driven methods.} A recent line of research has approached the problem of shape reconstruction in a data-driven manner by using a large dataset to learn a prior over the distribution of commonly occurring surfaces and how they correlate with the input point cloud.
These approaches are typically fast and robust to noisy inputs compared to non-data-driven approaches.
However, in such methods, the resulting reconstruction highly depends on the quality of such priors.

Several works have proposed to use a \emph{global} prior to capturing the distribution over full 3D object surfaces~\cite{Chen:2019:implicit, Mescheder:2019:Occupancy, park2019deepsdf}.
These methods define such a prior as a single latent vector representing the shape, which is then used as a condition in a fully connected network to decode the SDF of a given query point.
Usually, the decoder is trained on large data sets with a point-cloud encoder~\cite{Mescheder:2019:Occupancy, Chen:2019:implicit}.
However, Park et al.~\cite{park2019deepsdf} proposed to train the decoder directly on such data sets and then optimize the latent vector to match the noisy point cloud during inference.
Recently, Zhang et al.~\cite{zhang20233dshape2vecset} proposed to use richer global priors. 
They introduced an encoder-decoder network that encodes the input point cloud using attention modules into a set of latent vectors representing the shape, which are then used to predict the SDF for a set of query points using cross-attention modules.

Other works have opted to condition their models with \emph{local} priors.
Siddiqui et al.~\cite{siddiqui2021retrievalfuse} encoded the input point clouds in a set of latent scene patches. 
These latent vectors are used to query a database of latent vectors from patches obtained from the training set.
The obtained patches are then blended together using an attention mechanism.
Ma et al.~\cite{On-SurfacePriors2022} incorporated \emph{local} priors by including a network pre-trained on a large number of surface patches which classifies a point as being on the surface or not. 
This network is used to guide an optimization process that learns the shape's SDF using another neural network.
Jiang et al.~\cite{jiang2020lig} pre-trained an SDF encoder-decoder on a large data set of object parts.
Then, during the optimization process, only the latent codes of the different parts of the object are optimized.
Chen et al.~\cite{chen2022ndc} propose a dual contouring method learned on a small local prior.

Since \emph{global} and \emph{local} priors provide complementary information about the shape, a common approach is to use a prior in the \emph{medium} range using a hierarchical encoder-decoder network.
These approaches reduce the input point cloud to a simplified representation, e.g., voxelization or subsampled point cloud, which is then enriched by the global information provided by the bottleneck of the encoder-decoder architecture.
Chibane et al.~\cite{chibane20ifnet} and Peng et a.~\cite{Peng2020ECCV} proposed a 3DCNN encoder-decoder network to encode the sparse or noisy point cloud to later predict the SDF for an arbitrary point around the surface.
Chibane et al.~\cite{chibane2020ndf} extended this work to predict an unsigned distance field, which allowed them to represent complex open surfaces.
Tang et al.~\cite{tang2021sign} extended the work of Peng et al.~\cite{Peng2020ECCV} to include test-time optimization to improve out-of-distribution point clouds.
Ummenhofer and Koltun~\cite{Ummenhofer_2021_ICCV} proposed a CNN that works directly on an Octree, from which the model was able to predict the SDF.
Wang et al.~\cite{wang2022dualoctree} also represented the input point cloud with an octree, from which they constructed a graph. 
This graph was further processed by a GCN encoder-decoder to generate an embedding for each octree node, from where the final SDF is predicted.
Dai et al.~\cite{dai2020sgnn} instead used a 3D sparse encoder-decoder network to complete partial 3D scans and predict a complete SDF.
Lionar et al.~\cite{Lionar2021WACV} also developed an encoder-decoder network but used instead the projection of the input point cloud to a set of arbitrary 2D planes, from which the final SDF was predicted.
Boulch and Marlet~\cite{Boulch_2022_CVPR} recently proposed to use an encoder-decoder network that directly worked with points, avoiding discretization artifacts from voxel-based representations.
Although all these methods work relatively well when compared with methods that use \emph{global} or \emph{local} priors alone, they struggle to accurately capture fine local details of the shapes.

Erler et al.~\cite{Points2Surf2020} proposed to explicitly model \emph{global} and \emph{local} priors directly from point clouds using two different branches.
Each branch used a PointNet~\cite{qi2016pointnet} architecture, to process the local patch around the query point in the local branch, and a point cloud representing the complete shape in the global branch.
While the local branch was able to capture high-frequency details relatively well, they used a weak global prior due to the small subset of points selected to represent the shape. 
Our approach addresses the limitations of all these methods by incorporating strong \emph{global} and \emph{local} priors.

\begin{figure*}[t]
    \centering
    \includegraphics[width=\linewidth]{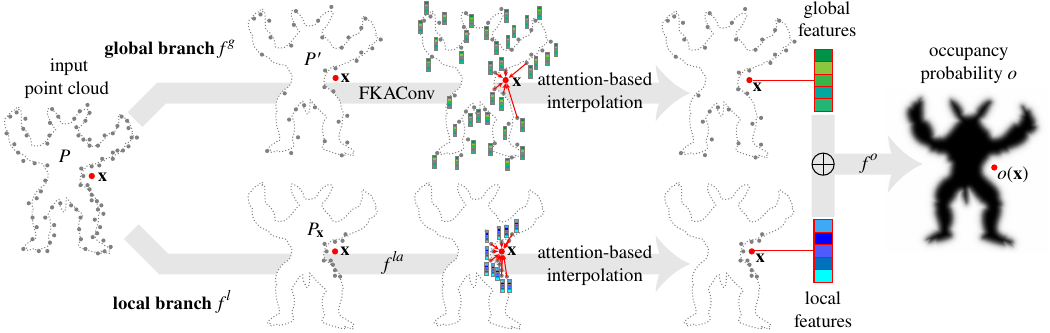}
    \caption{\name computes the occupancy probability at a query point $\mathbf{x}$ given a noisy point cloud $P$. A \emph{global} branch processes a sparse subset $P' \diff{\subseteq} P$ \diff{using} point convolutions, followed by an attention-based interpolation to get features at $\mathbf{x}$ that capture the coarse shape of the point cloud. A \emph{local} branch processes a local patch $P_\mathbf{x} \subset P$ using a PointNet~\cite{qi2016pointnet} with attention-based aggregation to get features at $\mathbf{x}$ that capture the detailed shape of the point cloud near $\mathbf{x}$. Global and local features are aggregated to compute the occupancy probability at $\mathbf{x}$.}
    \label{fig:architecture}
\end{figure*}

\section{Method}
\label{sec:method}

The goal of our method is to take as input an unoriented point cloud $P = \{\mathbf{p}_1, \mathbf{p}_2, \dots, \mathbf{p}_n \}$ that was sampled from an unknown watertight surface $\mathcal{S}^\text{gt}$ with a noisy sampling process, and output a surface $\mathcal{S}$ that approximates $\mathcal{S}^\text{gt}$ as closely as possible. Similar to several previous approaches, we define the surface $\mathcal{S}$ using an implicit representation, since this guarantees watertightness and naturally handles arbitrary surface topology in a smooth and differentiable way. More specifically, $\mathcal{S}$ is defined as the $0.5$-level set of an occupancy field $o(\mathbf{x})$:
$\mathcal{S} \coloneqq \{\mathbf{x}\ |\ o(\mathbf{x}) = 0.5\}$.

We train a network $f_\theta(\mathbf{x}, P)$ with parameters $\theta$ to model the field $o$ given a point cloud $P$:
\begin{equation}
    o(\mathbf{x}) \coloneqq f_\theta(\mathbf{x}, P)
\end{equation}
The network $f$ uses two branches: i) a \emph{global} branch $f^g(\mathbf{x}, P')$ that performs point convolutions~\cite{boulch2020fka} on a sparse random subset of points $P' \diff{\subseteq} P$ and effectively learns a global prior over the coarse shape of $\mathcal{S}$ given the input points $P$, and ii) a \emph{local} branch $f^l(\mathbf{x}, P_\mathbf{x})$ that processes a small local patch $P_\mathbf{x} \subset P$ around $\mathbf{x}$ and effectively learns a local prior over the detailed shape of local surface patches. 
Each branch outputs a feature vector for a given query point $\mathbf{x}$ that is combined into a single feature vector before being processed by a small MLP $f^o$ that outputs the occupancy probability $o(\mathbf{x})$:
\begin{equation}
    f_\theta(\mathbf{x}, P) \coloneqq f^o\big(f^g(\mathbf{x}, P') \oplus f^l(\mathbf{x}, P_\mathbf{x})\big),
\end{equation}
where $\oplus$ is the operation used to combine the two feature vectors, a sum in our experiments. Here, we omit the parameters of the networks $f^o$, $f^g$, and $f^l$ to avoid a cluttered notation. Figure~\ref{fig:architecture} illustrates our architecture.


In the following, we describe the architecture of \name, including the global and local branches in Section~\ref{sec:architecture}, followed by a description of the training and inference setups in Sections~\ref{sec:training} and~\ref{sec:inference}, respectively.

\subsection{Architecture}
\label{sec:architecture}

\paragraph*{Global Branch}
\label{sec:global}
The global branch $f^g(\mathbf{x}, P')$ takes as input a random subset $P' \diff{\subseteq} P$ and a 3D query point $\mathbf{x}$ and outputs a \emph{global} feature vector
for the point $\mathbf{x}$, which encodes information about the coarse shape of the point cloud. We implement the global branch using POCO~\cite{Boulch_2022_CVPR}, which consists of two main components: i) a point convolution module that computes a feature vector $\mathbf{z}'_i$ for each sparse point $\mathbf{p}'_i \in P'$, followed by ii) an interpolation module that interpolates the feature vectors $\mathbf{z}'_i$
to get the global feature vector at point $\mathbf{x}$.

The point convolution module uses FKAConv~\cite{boulch2020fka} to process the sparse point cloud $P'$ into a feature vector for each point:
\begin{equation}
    Z' = \text{FKAConv}(P'),
\end{equation}
where $Z' = \{\mathbf{z}'_1, \mathbf{z}'_2, \dots, \mathbf{z}'_{|P'|} \}$ is the set of feature vectors at each sparse point. Due to limitations both in performance and network capacity, convolutions can only be performed on the sparse subset $P'$ instead of the full point cloud $P$, with $|P'| = 10\text{k}$ in our experiments. This module consists of 10 layers of convolutions. Each layer uses a convolution kernel that operates over the 16 nearest neighbors of each point.

Given a query point $\mathbf{x}$, the interpolation module interpolates the feature vectors $\mathbf{z}'_i$ at the nearest neighbors $\mathcal{N}'_\mathbf{x}$ of the query point to get the global feature vector 
using an attention-based weighting:
\begin{align}
    \label{eq:gobal_interp}
    f^g(\mathbf{x}, P') \coloneqq f^{gb}\Big( \sum_{j \in \mathcal{N}'_\mathbf{x}} w_{\mathbf{x},j}\ f^{ga}\big((\mathbf{x} - \mathbf{p}'_j) \| \mathbf{z}'_j\big)\Big)\\
    \label{eq:gobal_interp_weight_attention}
    \text{with } w_{\mathbf{x},j} \coloneqq \frac{1}{k} \sum_{k=1}^{64} \text{softmax}_j\ f^{gw}_k\big((\mathbf{x} - \mathbf{p}'_j) \| \mathbf{z}'_j\big),
\end{align}
where $\|$ denotes concatenation, $f^{ga}$, $f^{gb}$ are two MLPs that transform the feature vectors before and after the weighted sum, and $f^{gw}_k$ are learned weighting functions, each implemented as a single linear layer. Analogous to the attention heads in multi-head attention, multiple different weighting functions are used as a form of ensemble learning, $64$ in our experiments.
Note that when evaluating multiple query points $\mathbf{x}$ for a point cloud, the point convolution module only needs to be evaluated once, while the interpolation module needs to be evaluated once per query point.

\paragraph*{Local Branch}
The local branch $f^l(\mathbf{x}, P_\mathbf{x})$ processes a local patch $P_\mathbf{x}$ around the query point $\mathbf{x}$ and outputs a \emph{local} feature vector for the point $\mathbf{x}$, which encodes information about the detailed shape of the point cloud near $\mathbf{x}$. We base the local branch on the popular PointNet~\cite{qi2016pointnet} architecture, which has been successfully applied in various methods that process local point cloud patches~\cite{GuerreroEtAl:PCPNet:EG:2018, rakotosaona2019pointcleannet}. We modify the architecture with an attention-based aggregation, instead of the original max- or sum-based aggregation, which we found to improve performance.

We define the local patch $P_\mathbf{x}$ as the $50$ nearest neighbors of the query point $\mathbf{x}$. We normalize the patch by centering it at the origin and scaling it to fit into a unit sphere, obtaining the normalized patch $\bar{P}_\mathbf{x}$. Subsequently, we apply PointNet with attention-based aggregation similar to Eqs.~\ref{eq:gobal_interp} and~\ref{eq:gobal_interp_weight_attention}, but without using multiple attention heads:

\begin{align}
    f^l(\mathbf{x}, P_\mathbf{x}) \coloneqq f^{lb}\big( \sum_{\bar{\mathbf{p}}_j \in \bar{P}_\mathbf{x}} v_{j}\ f^{la}(\bar{\mathbf{p}}_j)\big)\\
    \label{eq:local_interp_weight}
    \text{with } v_{j} \coloneqq \text{softmax}_j\ f^{lv}\big(f^{la}(\bar{\mathbf{p}}_j)\big),  
\end{align}
where $f^{lv}$ is a learned weighting function implemented as linear layer, and $f^{la}$, $f^{lb}$ are two MLPs that transform the feature vectors before and after the weighted aggregation.

\subsection{Training Setup}
\label{sec:training}

We train our network with a binary cross-entropy loss $\text{BCE}(o(\mathbf{x}), o^\text{gt}(\mathbf{x}))$ supervised by the ground-truth occupancy $o^\text{gt}(\mathbf{x})$ on query points defined by the Points2Surf \ABC var-noise training set~\cite{Points2Surf2020}.
We train with AdamW (lr=0.001, betas=(0.9, 0.999), eps=1e-5, weight\_decay=1e-2, amsgrad=False) for 150 epochs with scheduler steps at 75 and 125 epochs. On our training machine, we can fully utilize all 4 NVIDIA A40 GPUs with distributed data-parallel training using a total batch size of 50 and 48 workers. The other hyperparameters are mostly based on POCO, namely $10k$ manifold points, a network decoder $k$ of 64 and 2 output classes. One change is the increased latent size of 128, which was 32 in POCO. The additional hyperparameters for the local branch are a PointNet latent size of 256 and a patch size of 50. The training takes about 5 hours.



\subsection{Inference Setup}
\label{sec:inference}

We use the inference setup from POCO~\cite{Boulch_2022_CVPR}, which differs from the training setup in two main aspects: First, we perform test-time augmentation in our global branch to obtain more reliable results. Second, we sample query points in a grid and use a variant of marching cubes to reconstruct a mesh. We describe both in more detail below. 


\paragraph*{Test-time augmentation.} The sparse subsample $P' \diff{\subseteq} P$ used for the global branch may miss important geometric detail. To improve robustness, we compute the per-point feature vectors $\mathbf{z}'_i$ for multiple different random subsamples $P'_1, P'_2, \dots$, until each point in $P$ is included in at least $10$ subsamples. The $\geq 10$ different feature vectors for each point in $P$ are then averaged before performing the interpolation step.

\paragraph*{Mesh reconstruction.} We place query points in a $257^3$ grid and use a variant of marching cubes~\cite{lorensen1987marching} proposed in POCO to obtain a mesh from the occupancy field $o(\mathbf{x})$. That marching cubes variant uses a region-growing strategy starting from the input points to avoid the costly evaluation at all grid points, and super-samples marching-cube edges that intersect a surface to get a more accurate estimate of the intersection point.

\section{Results}
\label{sec:experiment}

We evaluate \name by comparing our surface reconstruction performance to several state-of-the-art methods, both data-driven and non-data-driven. We show both quantitative and qualitative comparisons in Section~\ref{sec:comparison}. Additionally, we provide an ablation to empirically validate our main design choices in Section~\ref{sec:ablation}.


\paragraph*{Metrics}

We use three well-known metrics to evaluate the error of our reconstructed surfaces: the Chamfer distance, the F1-score, and the normal error. We evaluate each metric at $100k$ random surface samples for the Chamfer distance and normal error, or volume samples for the IoU.
This results in roughly $\pm 0.5\%$ variance between different runs.


The \emph{Chamfer distance}~\cite{Barrow:1977:Chamfer,fan2017point} measures the distance between two point sets. We use it to measure the distance between reconstructed and GT surface samples. It is defined as:
\begin{equation}
\label{eq:chamfer}
\frac{1}{|A|} \sum_{\mathbf{p}_i \in A} \min_{\mathbf{p}_j \in B} \|\mathbf{p}_i - \mathbf{p}_j\|^2_2\ + \frac{1}{|B|} \sum_{\mathbf{p}_j \in B} \min_{\mathbf{p}_i \in A} \|\mathbf{p}_j - \mathbf{p}_i\|^2_2,
\end{equation}
where $A$ and $B$ are point sets of size $100k$ sampled on the surface of the GT object and the reconstructed object.






The \emph{F1 Score}~\cite{taha2015metrics} measures the overlap between the ground truth surface and the region enclosed by the reconstructed surface, similar to the IoU. It weights precision and recall equally.



The \emph{normal error} measures the difference between the normals of the reconstructed surface and the ground truth normals. We sample $100k$ points uniformly on the ground truth mesh $A$ and the reconstructed mesh $B$, storing the normals of their originating faces. Then, we find the closest neighbor of each point $b \in B$ in $A$. We report the average angle between the normals of these point pairs: $\frac{1}{n_s}\sum_{i=1}^{n_s}(\arccos(\mathbf{n}^A_i \cdot \mathbf{n}^B_i))$, where $\mathbf{n}^A_i$ and $\mathbf{n}^B_i$ are ground truth and reconstructed normals, respectively.


\paragraph*{Datasets}
We evaluate our method on the set of dataset variants introduced in P2S~\cite{Points2Surf2020}: 

\begin{figure}[t]
    \centering
    \includegraphics[width=\linewidth]{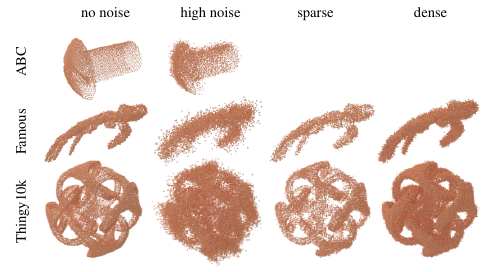}
    \caption{
    Point cloud examples of the data sets used in our evaluation.}
    \label{fig:datasets}
\end{figure}

\begin{itemize}
    \item The \ABC variant of P2S~\cite{Points2Surf2020} is a subset of the \ABC dataset by Koch et al.~\cite{Koch_2019_CVPR} and contains $4950$ points clouds from high-quality CAD meshes in the training set and $100$ point clouds in the test set.
    \item The \famous~\cite{Points2Surf2020} dataset consists of 22 diverse well-known meshes, including the Stanford Bunny, the Utah Teapot, and the Armadillo. We use this dataset for testing only.
    \item A subset of $100$ shapes from the \thingi~\cite{Thingi10K} dataset are used as additional test set. The \thingi dataset contains a variety of CAD shapes, but also more organic shapes like statues.
    \item The \real~\cite{Points2Surf2020} dataset consists of 3 real-world point clouds.
\end{itemize}

All synthetic point clouds were created with the simulated scanner BlenSor~\cite{gschwandtner2011blensor} with a scanner resolution of $176 \times 144$, using a random number of scans between $5$ and $30$. Each dataset comes in up to five variants:

\begin{itemize}
    \item \emph{no noise}: A version without noise
    \item \emph{med. noise}: A version with noise using a standard deviation of $0.01L$, where $L$ is the largest side of the object's bounding box.
    \item \emph{high noise}: A version with noise using a standard deviation of $0.05L$.
    \item \emph{var. noise}: A version with variable noise, where the amount of noise used for a given shape is sampled uniformly in $[0, 0.05L]$ and the number of scans in $[5, 30]$.
    \item \emph{sparse}: A version with medium noise where all shapes only uses $5$ scans, resulting in point clouds between $2k$ and $22k$ points.
    \item \emph{dense}: A version with medium noise where all shapes use $30$ scans, resulting in point clouds between $5k$ and $112k$ points.
\end{itemize}

For a fair comparison, we train all data-driven methods on the \ABC var. noise dataset and evaluate them with each test set.
Some point cloud examples of these data sets are illustrated in Figure~\ref{fig:datasets}.


\subsection{Comparisons}
\label{sec:comparison}


We compare \name to several recent data-driven and non-data-driven reconstruction methods. PGR~\cite{lin2023gaussformula}, Neural-IMLS (IMLS)~\cite{wang2022NeuralIMLS} and Shape as Points (SAP-O) are non-data-driven methods that do not train on a large dataset and instead directly fit a surface to the input point cloud. Shape as Points also has a data-driven variant (SAP) that uses a trained network. Additionally, we use Points2Surf (P2S)~\cite{Points2Surf2020} and POCO~\cite{Boulch_2022_CVPR} as data-driven methods.
%
%
We took the best available variants and settings for each method: 
For PGR, we use the default parameters \textit{wmin=0.0015, alpha=1.05} for no noise, med noise and var. noise. We use the following adapted parameters for the other datasets: \textit{wmin=0.03, alpha=2.0} for high noise, \textit{wmin=0.03, alpha=1.5} for dense and sparse.
We use \textit{thingi-noisy} for SAP-O, 
\textit{vanilla} for P2S
, and \textit{10k-FKAConv-InterpAttentionKHeadsNet} for POCO. 
We used the provided \textit{noise-large} configuration for SAP. 
For IMLS, we used the results provided by the authors (high noise datasets were not provided by the authors). Note that IMLS was developed concurrently with our work.


\paragraph*{Qualitative Comparison}

\begin{figure*}[p]
    \centering
    \includegraphics[width=\linewidth]{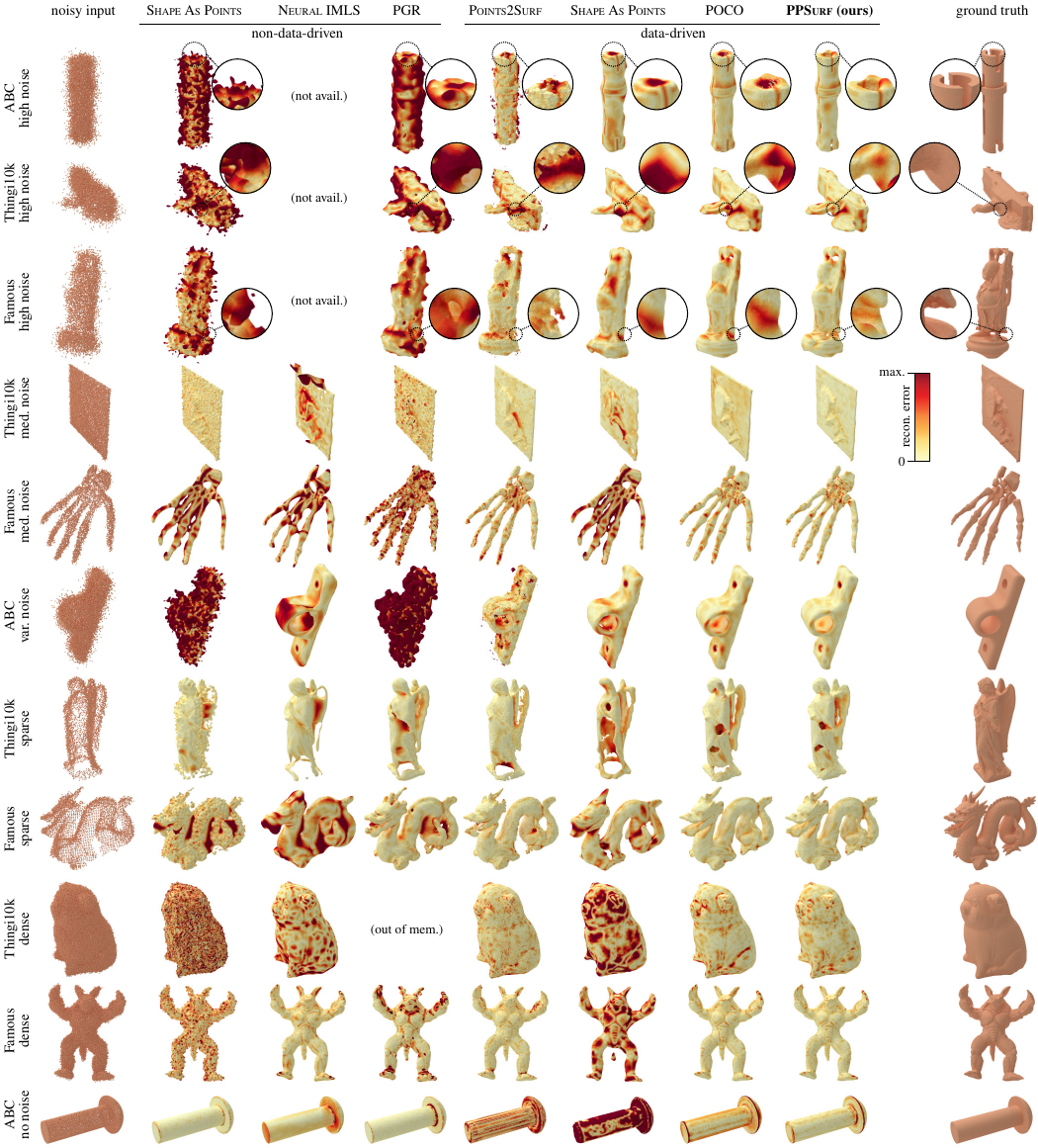}
    \caption{Qualitative comparison to all baselines. We evaluate one example from each dataset variant (except for the no-noise variants, where we only show one example due to space constraints). Colors show the distance of the reconstructed surface to the ground-truth surface. Due to our combined local and global branches, \name reconstructs details more accurately than the baselines, especially in the presence of strong input noise. Note that results for Neural IMLS are not provided by the authors for the high-noise dataset variants. See the supplementary material for a qualitative comparison on all shapes in our test sets.}
    \label{fig:qual_comp}
\end{figure*}

Figure~\ref{fig:qual_comp} shows comparisons for one example of each dataset variant. While non-data-driven methods give competitive results on low-noise results, \name has a clear advantage with sparse and noisy point clouds.

\begin{figure*}[t]
    \centering
    \includegraphics[width=\linewidth]{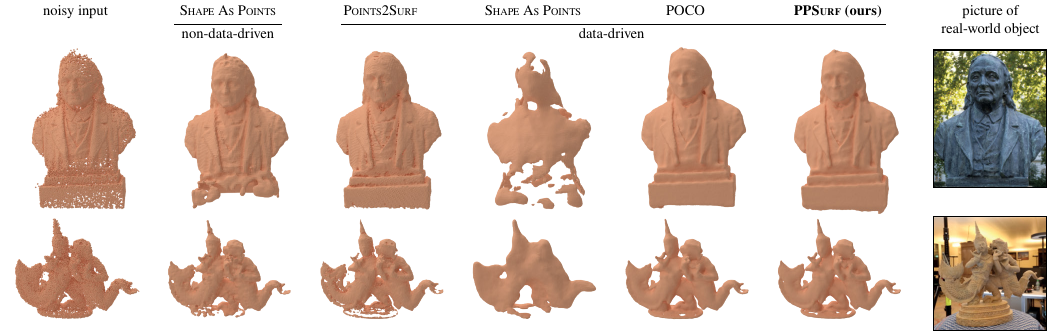}
    \caption{Real-world reconstructions. We compare to all baselines on the two point clouds that were obtained from real-world objects.}
    \label{fig:qual_comp_real}
\end{figure*}

We show examples on real-world point clouds in Figure~\ref{fig:qual_comp_real}, where \name produces clearer edges and finer details.

\paragraph*{Quantitative Comparison}

Table~\ref{tab:quant_comparison} shows the performance of \name on all dataset variants. We report the average over all shapes in the test set. Similar to the qualitative results, POCO, \name and the non-data-driven methods share the first place in most low-noise dataset variants, but \name 50NN takes the lead in almost all other dataset variants.
This confirms that adding the local branch does indeed improve the local reconstruction.

\begin{table*}[t]

\caption{Comparison of reconstruction errors. We show the Chamfer distance, F1 Score and normal error between reconstructed and ground-truth surfaces averaged over all shapes in a dataset. Apart from a few noise-free datsets, \name consistently performs similar or better than the baselines.
Note that the mean performance of Neural IMLS does not include results of the high noise datasets, which are likely to favour \name. Due to out-of-memory errors, PGR could not reconstruct all shapes, which are ignored here.
Best results per row are marked in bold and the second-best results are underlined.}

\centering 
\scriptsize
\setlength{\tabcolsep}{2.5pt}

\begin{tabular}{lcccccccccccccccccccccccc}
\toprule
    Dataset & \multicolumn{7}{c}{Chamfer Distance (x100)~$\Downarrow$} & \multicolumn{7}{c}{F1~$\Uparrow$} & \multicolumn{7}{c}{Normal Error~$\Uparrow$} \\
    \cmidrule(lr){2-8} \cmidrule(lr){9-15} \cmidrule(lr){16-22} 
    &
    IMLS & PGR & SAP-O &  SAP &  P2S & POCO  & \name &
    IMLS & PGR & SAP-O &  SAP &  P2S & POCO  & \name &
    IMLS & PGR & SAP-O &  SAP &  P2S & POCO  & \name \\
\midrule
    ABC var. noise &
    1.08 & 1.60 & 1.18 & 1.18 & 0.84 &  \underline{0.70} &   \textbf{0.66} &
    0.78 & 0.50 & 0.67 & 0.79 & 0.83 &  \underline{0.89} &   \textbf{0.90} &
    0.55 & 1.29 & 1.11 & 0.52 & 0.65 &  \underline{0.32} &   \textbf{0.30} \\
\midrule
    ABC no noise &
    \textbf{0.48} & 0.53 & 0.63 & 1.08 & 0.61 &  \underline{0.50} & \textbf{0.48} &
    \underline{0.92} & \underline{0.92} & 0.88 & 0.80 & 0.88 &  \textbf{0.94} &   \textbf{0.94} &
    \underline{0.20} & 0.26 & 0.30 & 0.51 & 0.31 &  \textbf{0.19} &   \textbf{0.19} \\
    
    Famous no noise &
    \textbf{0.35} & \underline{0.36} & \textbf{0.35} & 0.99 & 0.46 &  0.39 &  0.37 &
    \underline{0.95} & \underline{0.95} & \underline{0.95} & 0.84 & 0.93 &  \underline{0.95} &   \textbf{0.96} &
    \textbf{0.44} & 0.48 & \textbf{0.44} & 0.75 & 0.57 &  \underline{0.46} &   \underline{0.46} \\
    
    Thingi10k no noise &
    0.40 & \textbf{0.30} & 0.43 & 0.89 & 0.39 &  \underline{0.33} &   \underline{0.33} &
    0.93 & \underline{0.95} & 0.92 & 0.85 & 0.93 &  \underline{0.95} &   \textbf{0.96} &
    0.25 & \underline{0.19} & 0.23 & 0.49 & 0.32 &  \textbf{0.18} &   \underline{0.19} \\
    
    \textbf{Mean no noise} &
    0.41 & \underline{0.40} & 0.47 & 0.99 & 0.49 & 0.41 & \textbf{0.39} &
    0.93 & \underline{0.94} & 0.92 & 0.83 & 0.91 & \textbf{0.95} & \textbf{0.95} &
    \underline{0.30} & 0.31 & 0.33 & 0.58 & 0.40 & \textbf{0.28} & \textbf{0.28} \\
\midrule  
    Famous med. noise &
    0.54 & 0.95 & 0.58 & 1.06 & 0.52 &  \underline{0.49} &   \textbf{0.48} &
    0.91 & 0.60 & 0.90 & 0.83 & 0.92 &  \textbf{0.94} &   \underline{0.93} &
    0.57 & 1.35 & 0.91 & 0.78 & 0.63 &  \underline{0.53} &   \textbf{0.54} \\
    
    Thingi10k med. noise &
    0.58 & 0.93 & 0.56 & 0.93 & 0.44 &  \underline{0.39} &   \textbf{0.38} &
    0.90 & 0.57 & 0.89 & 0.85 & \underline{0.92} &  \textbf{0.94} &   \textbf{0.94} &
    0.37 & 1.32 & 0.78 & 0.50 & 0.38 &  \textbf{0.24} &   \underline{0.25} \\
    
    \textbf{Mean med. noise} &
    0.56 & 0.94 & 0.57 & 0.99 & 0.48 & \underline{0.44} & \textbf{0.43} &
    0.91 & 0.59 & 0.89 & 0.84 & \underline{0.92} & \textbf{0.94} & \textbf{0.94} &
    0.47 & 1.33 & 0.85 & 0.64 & 0.50 & \textbf{0.38} & \underline{0.39} \\
\midrule
    ABC high noise &
    -- & 1.90 & 1.96 & 1.51 & 1.24 &  \underline{1.00} &   \textbf{0.97} &
    -- & 0.42 & 0.49 & 0.75 & 0.78 &  \underline{0.84} &   \textbf{0.85} &
    -- & 1.35 & 1.42 & 0.65 & 0.99 &  \underline{0.43} &   \textbf{0.41} \\
    
    Famous high noise &
    -- & 1.86 & 1.80 & 1.62 & 1.14 &  \underline{1.11} &   \textbf{1.01} &
    -- & 0.50 & 0.59 & 0.78 & \underline{0.84} &  \underline{0.84} &   \textbf{0.85} &
    -- & 1.35 & 1.39 & 0.91 & 1.04 &  \underline{0.76} &   \textbf{0.72} \\
    
    Thingi10k high noise &
    -- & 1.94 & 1.89 & 1.45 & 1.08 &  \underline{0.92} &   \textbf{0.83} &
    -- & 0.51 & 0.60 & 0.80 & 0.84 &  \underline{0.87} &   \textbf{0.88} &
    -- & 1.31 & 1.36 & 0.64 & 0.90 &  \underline{0.47} &   \textbf{0.43} \\
    
    \textbf{Mean high noise} &
    -- & 1.90 & 1.88 & 1.53 & 1.16 & \underline{1.01} & \textbf{0.94} &
    -- & 0.32 & 0.56 & 0.78 & 0.82 & \underline{0.85} & \textbf{0.86} &
    -- & 1.34 & 1.39 & 0.73 & 0.98 & \underline{0.55} & \textbf{0.52} \\
\midrule  
    Famous sparse &
    0.90 & 0.88 & 0.71 & 1.24 & 0.77 &  \underline{0.67} &   \textbf{0.64} &
    0.86 & 0.88 & 0.88 & 0.74 & \underline{0.89} &  \textbf{0.92} &   \textbf{0.92} &
    0.68 & 0.75 & 0.86 & 0.89 & 0.71 &  \underline{0.60} &   \textbf{0.61} \\
    
    Thingi10k sparse &
    0.82 & 0.89 & 0.86 & 1.35 & \underline{0.78} &  \textbf{0.63} &   \textbf{0.63} &
    0.85 & 0.86 & 0.84 & 0.73 & \underline{0.87} &  \textbf{0.90} &   \textbf{0.90} &
    0.48 & 0.53 & 0.76 & 0.73 & 0.51 &  \textbf{0.37} &   \underline{0.39} \\
    
    \textbf{Mean sparse} &
    0.86 & 0.88 & 0.79 & 1.29 & 0.77 & \underline{0.65} & \textbf{0.63} &
    0.86 & 0.87 & 0.86 & 0.73 & \underline{0.88} & \textbf{0.91} & \textbf{0.91} &
    0.58 & 0.64 & 0.81 & 0.81 & 0.61 & \textbf{0.48} & \underline{0.50} \\
\midrule  
    Famous dense &
    0.45 & 0.70 & 0.53 & 0.96 & \underline{0.41} &  0.42 &   \textbf{0.40} &
    0.93 & 0.43 & 0.90 & 0.86 & \underline{0.94} &  \textbf{0.95} &   \textbf{0.95} &
    0.52 & 1.33 & 1.00 & 0.74 & 0.59 &  \underline{0.49} &   \textbf{0.48} \\
    
    Thingi10k dense &
    0.49 & 0.67 & 0.54 & 0.88 & 0.36 &  \underline{0.35} &   \textbf{0.33} &
    0.91 & 0.47 & 0.89 & 0.87 & 0.94 &  \underline{0.95} &   \textbf{0.96} &
    \underline{0.30} & 1.23 & 0.84 & 0.47 & 0.33 &  \textbf{0.21} &   \textbf{0.21} \\
    
    \textbf{Mean dense} &
    0.47 & 0.69 & 0.53 & 0.92 & \underline{0.39} & \underline{0.39} & \textbf{0.37} &
    0.92 & 0.45 & 0.89 & 0.86 & 0.94 & \underline{0.95} & \textbf{0.96} &
    0.41 & 1.28 & 0.92 & 0.60 & 0.46 & \underline{0.35} & \textbf{0.34} \\
\midrule  
    \textbf{Mean overall} &
    \underline{0.61} & 1.04 & 0.93 & 1.16 & 0.70 &  \underline{0.61} &   \textbf{0.58} &
    0.89 & 0.66 & 0.80 & 0.81 & 0.89 &  \underline{0.91} &   \textbf{0.92} &
    \underline{0.43} & 0.98 & 0.88 & 0.66 & 0.61 &  \textbf{0.40} &   \textbf{0.40} \\
\bottomrule
\end{tabular}

\label{tab:quant_comparison}
\end{table*}

\paragraph*{Computation Time and Memory Consumption}

Training \name on the \ABC var-noise training set was done in 5 hours on 4 NVIDIA A40 GPUs and 48 AMD EPYC-Milan cores. 
We reconstruct all shapes in our test sets on a single A40 and 48 CPU cores.
See the timings and memory consumption in Table~\ref{tab:resources}. While non-data-driven methods tend to be faster than data-driven ones, SAP is a lightning-fast exception. \name with small patch sizes has a negligible impact on resources compared to POCO.
Neural IMLS does not report timings. As it is concurrent work, we could not do our own measurements. While it is fast, PGR's memory usage varies a lot with point cloud size, between a few GB to going out-of-memory with >46GB on 21 shapes.




\begin{table}[t]

\caption{Comparison of reconstruction times and memory usage. We show the mean reconstruction time per shape and the maximum GPU-memory consumption for each method on the ABC var noise dataset.
200NN uses reconstruction batch size $25k$ instead of $50k$. PGR went out of memory on 21 shapes.}

\centering 
\footnotesize
\setlength{\tabcolsep}{2.5pt}

\begin{tabular}{cccc}
\toprule
             & Time per Shape    & Max GPU Memory    \\
\midrule
PGR          & 1.9 min           & >46GB        \\ 
SAP-O        & 1.1 min           & 3.8GB        \\ 
SAP          & 0.8sec            & 3.1GB    \\
P2S          & 13.5min           & 14.3GB   \\ 
POCO         & 1.6min            & 9.0GB    \\
\name 10NN   & 1.6min            & 9.1GB    \\ 
\name 25NN   & 1.7min            & 9.1GB    \\
\name 50NN   & 1.9min            & 9.3GB    \\
\name 100NN  & 2.6min            & 13.7GB   \\
\name 200NN  & 3.5min            & 13.2GB   \\
\bottomrule
\end{tabular}

\label{tab:resources}
\end{table}

\paragraph*{Discussion}

For dense and noise-free point clouds, non-data-driven methods such as PGR, SAP-O and especially IMLS are a good option. However, their performance is limited in the presence of typical point-cloud artifacts, due to missing data-driven priors. Data-driven methods such as SAP, P2S, POCO and \name can better deal with such artifacts. SAP is the fastest method but lacks accuracy, possibly due to its very small network. A bigger version could perhaps produce competitive results but would require non-trivial changes to the method.

P2S employs a relatively simple PointNet for global shape encoding, which results in a weak global prior that can not reach the quality of a more efficient encoder such as FKAConv. Furthermore, it reconstructs noisy surfaces, which is reflected in the relatively high normal error, even with noise-free inputs.

Apart from some noise-free datasets, only POCO is close to \name's quality. \name achieves similar results on low-noise point clouds, but significantly better reconstructions for noisy point clouds. When predicting the occupancy at the query points, POCO has no direct access to the full point cloud, only to a coarse latent representation. This inability to accurately represent local information is likely the reason why POCO tends to produce blobby structures and over-smooth the reconstructed surfaces. We avoid this by providing a latent code that captures local detail more accurately
by adding a local branch that directly encodes dense local patches of the point cloud.


\subsection{Ablation}
\label{sec:ablation}

We investigate several design choices in an ablation study on the \ABC var-noise test set. Most importantly, Table~\ref{tab:ablation_branches} shows that having both global and local branches gains a major advantage. Referring to Table~\ref{tab:ablation_patchsize}, the optimal local patch size lies in the range of $25NN$ to $100NN$. Further, attention is a better symmetric operation than max, and concatenating features is similar to summing them. This can be seen in Table~\ref{tab:ablation_misc}. 
Please see the supplementary for an evaluation of the most relevant variants on all datasets.
We compare the following variants of our method:
\begin{itemize}
    \item \emph{Full} is the full method as described in Section~\ref{sec:method}.
    \item For \emph{Only Local}, we set the global features to zeros, disabling this branch. Based on the results of this experiment, we conclude that this model can not reliably encode any surface since it lacks global knowledge of the surface to reconstruct. 
    \item \emph{Only Global} is similar to POCO as it omits the local branch. The results show that a global prior can help to obtain reliable reconstructions but with lower performance due to the missing fine details.
    \item For \emph{Sym Max}, we replace the attention-based interpolation used in the local branch with the max, effectively making this branch a PointNet~\cite{qi2016pointnet}. The results show an advantage for attention.
    \item In \emph{Merge Cat}, we concatenate the features of both branches instead of summing them, which leads to twice the input size for the final MLP. Results show that this is slightly worse than \emph{Full}.
    \item The \emph{QPoints} variant is the same as \emph{Merge Cat}, but additionally, we concatenate query point coordinates to the input of the learned weighting function $f^{lv}$. However, this results in a slightly worse performance than \emph{Full} and even \emph{Merge Cat}. 
    \item For the \emph{xNN} variants, we take the $x$ nearest neighbors for local subsample. \emph{Full} is equal to 50NN.
\end{itemize}

\begin{table}[t]

\caption{Branch Ablation Study. Using the ABC var-noise test set, we compare \name \emph{Full} to variants with disabled branches. The only-local variant failed to produce some meshes, which are ignored in the metrics. The best results per column are marked in bold.}
\centering 

\footnotesize
\setlength{\tabcolsep}{4pt}

\begin{tabular}{lcccccccccccc}
\toprule
    Model & Chamfer (x100)~$\Downarrow$ & F1 Score~$\Uparrow$ & Normal Error~$\Downarrow$ \\
\midrule
Only Local         & 2.69          & 0.36          & 1.56          \\
Only Global        & 0.70          & 0.89          & 0.33          \\
\textbf{\name Full} & \textbf{0.66} & \textbf{0.90} & \textbf{0.30} \\
\bottomrule
\end{tabular}

\label{tab:ablation_branches}
\end{table}

\begin{table}[t]

\caption{Patch Size Ablation Study. Using the ABC var-noise test set, we compare \name \emph{Full} (which is 50NN) to variants with different patch sizes. The best results per column are marked in bold.}
\centering 

\footnotesize
\setlength{\tabcolsep}{4pt}

\begin{tabular}{lcccccccccccc}
\toprule
    Model & Chamfer (x100)~$\Downarrow$ & F1 Score~$\Uparrow$ & Normal Error~$\Downarrow$ \\
\midrule
\name 10NN                  & 1.10          & \textbf{0.90} & 0.40          \\
\name 25NN                  & \textbf{0.66} & \textbf{0.90} & 0.31          \\
\textbf{\name Full}  & \textbf{0.66} & \textbf{0.90} & \textbf{0.30} \\
\name 100NN                 & \textbf{0.66} & \textbf{0.90} & \textbf{0.30} \\
\name 200NN                 & 0.67          & 0.89          & 0.31          \\
\bottomrule
\end{tabular}

\label{tab:ablation_patchsize}
\end{table}

\begin{table}[t]

\caption{Miscellanous Ablation Study. Using the ABC var-noise test set, we compare \name \emph{Full} (which uses Merge Sum and Sym Att) to more variants. The best results per column are marked in bold.}
\centering 

\footnotesize
\setlength{\tabcolsep}{4pt}

\begin{tabular}{lcccccccccccc}
\toprule
    Model & Chamfer (x100)~$\Downarrow$ & F1 Score~$\Uparrow$ & Normal Error~$\Downarrow$ \\
\midrule
\name Sym Max       & 1.11                                 & \textbf{0.90}       & 0.40                      \\
\name QPoints       & 0.67                                 & 0.89                & 0.31                      \\
\name Merge Cat     & \textbf{0.66}                        & \textbf{0.90}       & \textbf{0.30}             \\
\textbf{\name Full} & \textbf{0.66}                        & \textbf{0.90}       & \textbf{0.30}             \\
\bottomrule
\end{tabular}

\label{tab:ablation_misc}
\end{table}

\subsection{Limitations}

\begin{figure}[t]
    \centering
    \includegraphics[width=0.95\linewidth]{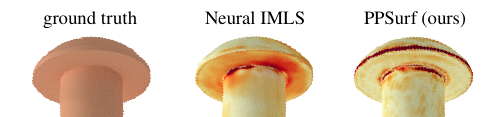}
    \caption{
    Limitations. Our method has difficulties to recover the edges of clean point clouds due to training with noisy point clouds.}
    \label{fig:limitations-2}
\end{figure}

\begin{figure}[t]
    \centering
    \includegraphics[width=0.95\linewidth]{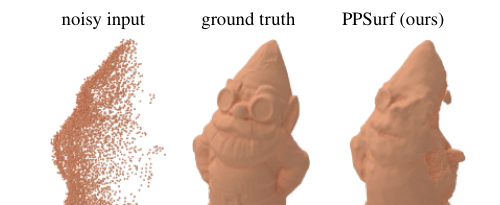}
    \caption{
    Limitations. Our method struggles with reconstructions of large missing areas in the input point cloud since we did not incorporate any generative model capabilities.}
    \label{fig:limitations}
\end{figure}

Reconstruction times are still non-interactive, due to the need to evaluate the occupancy at a large number of samples.
Possibilities for speed-ups include
more efficient sampling strategies to use fewer query points. 

As our learned priors were trained on noisy data to make \name more robust to noise, they also bias the reconstructed surface to some extent towards the distributions learned by the priors. This results in some loss of accuracy when applied to noise-free point clouds compared to some of the non-data-driven methods (see Figure~\ref{fig:limitations-2}). Learning a prior that is specialized to noise-free point clouds, or including more noise-free point clouds in our training set would alleviate this issue.

While \name is better than the baselines in filling scan shadows, it is not a generative method and cannot generate new geometric detail in large missing regions. This limits the size of missing regions that can be filled with plausible geometry. Combining \name with a generative model would be an interesting direction for future work. See Figure~\ref{fig:limitations} for an example of inaccurately filled scan shadows.

\section{Conclusion}

In this paper, we have introduced \name as a method for surface reconstruction from raw, unoriented point clouds.
In contrast to previous methods, \name incorporates strong local and global priors learned from data.
Whilst our global prior is based on a point convolutional neural network that processes the point cloud as a whole, fine details are preserved through the local prior based on dense local point cloud patches.
We have shown in extensive studies that \name is able to achieve better surface reconstructions than previous data-driven and non-data-driven methods, being more robust to noise in the input point cloud and preserving fine details at the same time.

In the future, we would like to investigate how modern techniques borrowed from generative models could improve the obtained reconstruction from sparse point clouds where large parts of the shape are missing.

\section{Acknowledgements}

This work has been supported by the FWF projects P24600-N23 and P32418-N31, the WWTF project ICT19-009 and the EU MSCA-ITN project EVOCATION (grant agreement 813170).

\bibliographystyle{eg-alpha-doi}
\bibliography{main}

\end{document}



\maketitle




\section{Supplementary}
\label{sec:supplementary}

The \emph{Intersection over Union} measures the relative amount of overlap between two volumetric regions. We use it to measure the relative overlap between the region enclosed by the ground truth surface $A_V$ and the region enclosed by the reconstructed surface $B_V$: $(A_V \cap B_V) / (A_V \cup B_V)$. We use a Monte-Carlo estimate of this IoU with $100$k sample points in the volume of the unit cube.

\begin{table}[t]

\caption{Comparison of IoU. We show the IoU between reconstructed and ground-truth surfaces averaged over all shapes in a dataset. \name performs similar or better than the baselines.
Best results per row are marked in bold and the second-best results are underlined.}

\centering 
\footnotesize
\setlength{\tabcolsep}{2pt}

\begin{tabular}{lccccccc}
\toprule
Dataset~\textbackslash~IoU~$\Uparrow$ & IMLS & PGR  & SAP-O        & SAP  & P2S              & POCO             & \name         \\ 
\midrule
ABC var. noise           & 0.69              & 0.37 & 0.56          & 0.69 & 0.75             & \underline{0.82} & \textbf{0.84} \\ 
\midrule
ABC no noise             & \underline{0.88}  & \underline{0.88} & 0.83          & 0.69 & 0.83             & \textbf{0.90}    & \textbf{0.90} \\
Famous no noise          & \textbf{0.92}     & 0.91 & \textbf{0.92} & 0.75 & \underline{0.88} & \textbf{0.92}    & \textbf{0.92} \\
Thingi10k no noise       & 0.89              & \underline{0.93} & 0.89          & 0.77 & 0.90             & \underline{0.93} & \textbf{0.94} \\
\textbf{Mean no noise}   & 0.89  & \underline{0.91} & 0.88          & 0.74 & 0.87             & \textbf{0.92}    & \textbf{0.92} \\ 
\midrule
Famous med. noise        & \underline{0.86}  & 0.44 & 0.83          & 0.74 & \underline{0.86} & \textbf{0.89}    & \textbf{0.89} \\
Thingi10k med. noise     & 0.84              & 0.42 & 0.83          & 0.77 & \underline{0.89} & \textbf{0.91}    & \textbf{0.91} \\
\textbf{Mean med. noise} & 0.85              & 0.43 & 0.83          & 0.75 & \underline{0.87} & \textbf{0.90}    & \textbf{0.90} \\ 
\midrule
ABC high noise           & --                & 0.28 & 0.35          & 0.63 & 0.67             & \underline{0.75} & \textbf{0.76} \\
Famous high noise        & --                & 0.34 & 0.44          & 0.65 & \underline{0.74} & \underline{0.74} & \textbf{0.76} \\
Thingi10k high noise     & --                & 0.35 & 0.45          & 0.70 & 0.76             & \underline{0.79} & \textbf{0.81} \\
\textbf{Mean high noise} & --                & 0.32 & 0.41          & 0.66 & 0.72             & \underline{0.76} & \textbf{0.78} \\ 
\midrule
Famous sparse            & 0.78              & 0.79 & 0.80          & 0.64 & 0.81             & \underline{0.85} & \textbf{0.86} \\
Thingi10k sparse         & 0.78              & 0.79 & 0.77          & 0.62 & 0.81             & \underline{0.84} & \textbf{0.85} \\
\textbf{Mean sparse}     & 0.78              & 0.79 & 0.79          & 0.63 & \underline{0.81} & \textbf{0.85}    & \textbf{0.85} \\ 
\midrule
Famous dense             & 0.88              & 0.28 & 0.83          & 0.77 & 0.90             & \underline{0.91} & \textbf{0.92} \\
Thingi10k dense          & 0.87              & 0.32 & 0.83          & 0.79 & \underline{0.91} & \textbf{0.93}    & \textbf{0.93} \\
\textbf{Mean dense}      & 0.87              & 0.3  & 0.83          & 0.78 & 0.90             & \underline{0.92} & \textbf{0.93} \\ 
\midrule
\textbf{Mean overall}    & 0.84              & 0.55 & 0.72          & 0.71 & 0.82             & \underline{0.86} & \textbf{0.87} \\
\bottomrule
\end{tabular}

\label{tab:quant_comparison_iou}
\end{table}

\begin{table*}[t]

\caption{All Datasets Patch Size Ablation Study. We compare the most relevant patch sizes on all datasets. The best results per row are marked in bold.}

\centering 
\footnotesize
\setlength{\tabcolsep}{1.5pt}

\begin{tabular}{lcccccccccccc}
\toprule
Dataset & \multicolumn{3}{c}{Chamfer (x100)~$\Downarrow$} & \multicolumn{3}{c}{IoU~$\Uparrow$} & \multicolumn{3}{c}{F1~$\Uparrow$} & \multicolumn{3}{c}{Normal Error~$\Downarrow$} \\
 \cmidrule(lr){2-4} \cmidrule(lr){5-7} \cmidrule(lr){8-10} \cmidrule(lr){11-13}
    \name & 25NN & 50NN & 100NN & 25NN & 50NN & 100NN & 25NN & 50NN & 100NN & 25NN & 50NN & 100NN \\
\midrule
ABC var. noise           & 0.664          & 0.660          & \textbf{0.659} & 0.835          & \textbf{0.838} & 0.835          & 0.898          & \textbf{0.901} & 0.899          & 0.306          & \textbf{0.297} & 0.301          \\ 
\midrule   
ABC no noise             & 0.486          & \textbf{0.482} & 0.497          & 0.902          & 0.902          & \textbf{0.903} & \textbf{0.941} & \textbf{0.941} & 0.940          & \textbf{0.190} & 0.194          & 0.199          \\
Famous no noise          & 0.396          & \textbf{0.373} & 0.419          & 0.920          & \textbf{0.923} & 0.906          & 0.955          & \textbf{0.958} & 0.945          & 0.460          & \textbf{0.455} & 0.491          \\
Thingi10k no noise       & \textbf{0.314} & 0.327          & 0.319          & \textbf{0.936} & \textbf{0.936} & \textbf{0.936} & \textbf{0.958} & \textbf{0.958} & \textbf{0.958} & \textbf{0.190} & \textbf{0.190} & \textbf{0.190} \\
\textbf{Mean no noise}   & 0.399          & \textbf{0.394} & 0.412          & 0.919          & \textbf{0.920} & 0.915          & 0.951          & \textbf{0.952} & 0.948          & 0.280          & \textbf{0.279} & 0.294          \\ 
\midrule   
Famous med. noise        & 0.496          & \textbf{0.484} & 0.485          & 0.883          & \textbf{0.887} & 0.883          & 0.934          & \textbf{0.935} & 0.933          & \textbf{0.539} & 0.543          & 0.542          \\
Thingi10k med. noise     & 0.384          & \textbf{0.383} & 0.384          & 0.910          & 0.910          & \textbf{0.912} & 0.943          & 0.943          & \textbf{0.944} & 0.245          & 0.245          & \textbf{0.244} \\
\textbf{Mean med. noise} & 0.440          & \textbf{0.434} & \textbf{0.434} & 0.896          & \textbf{0.898} & 0.897          & 0.938          & \textbf{0.939} & \textbf{0.939} & \textbf{0.392} & 0.394          & 0.393          \\ 
\midrule   
ABC high noise           & \textbf{0.953} & 0.968          & 0.972          & \textbf{0.763} & 0.759          & 0.759          & \textbf{0.850} & 0.846          & 0.847          & \textbf{0.399} & 0.409          & 0.406          \\
Famous high noise        & 1.002          & 1.010          & \textbf{0.992} & \textbf{0.759} & 0.757          & 0.749          & \textbf{0.853} & 0.852          & 0.845          & 0.725          & 0.723          & \textbf{0.701} \\
Thingi10k high noise     & \textbf{0.818} & 0.826          & 0.824          & \textbf{0.813} & 0.810          & 0.812          & \textbf{0.883} & 0.881          & \textbf{0.883} & \textbf{0.422} & 0.427          & 0.423          \\
\textbf{Mean high noise} & \textbf{0.925} & 0.935          & 0.929          & \textbf{0.779} & 0.775          & 0.773          & \textbf{0.862} & 0.860          & 0.858          & 0.515          & 0.520          & \textbf{0.510} \\
\midrule  
Famous sparse            & 0.673          & 0.636          & \textbf{0.629} & 0.852          & \textbf{0.857} & 0.853          & 0.914          & \textbf{0.918} & 0.915          & 0.616          & 0.606          & \textbf{0.600} \\
Thingi10k sparse         & 0.647          & \textbf{0.629} & 0.661          & \textbf{0.849} & 0.846          & \textbf{0.849} & 0.903          & 0.901          & \textbf{0.904} & 0.387          & 0.388          & \textbf{0.383} \\
\textbf{Mean sparse}     & 0.660          & \textbf{0.633} & 0.645          & 0.850          & \textbf{0.852} & 0.851          & \textbf{0.909} & \textbf{0.909} & \textbf{0.909} & 0.502          & 0.497          & \textbf{0.491} \\ 
\midrule  
Famous dense             & 0.411          & \textbf{0.403} & 0.404          & 0.910          & \textbf{0.918} & 0.908          & 0.949          & \textbf{0.954} & 0.947          & 0.482          & \textbf{0.476} & 0.488          \\
Thingi10k dense          & 0.332          & 0.327          & \textbf{0.326} & 0.930          & 0.932          & \textbf{0.933} & 0.957          & 0.958          & \textbf{0.959} & 0.210          & \textbf{0.209} & 0.212          \\
\textbf{Mean dense}      & 0.371          & \textbf{0.365} & \textbf{0.365} & 0.920          & \textbf{0.925} & 0.921          & 0.953          & \textbf{0.956} & 0.953          & 0.346          & \textbf{0.342} & 0.350          \\
\midrule   
\textbf{Mean overall}    & 0.583          & \textbf{0.578} & 0.582          & 0.866          & \textbf{0.867} & 0.864          & 0.918          & \textbf{0.919} & 0.917          & 0.398          & \textbf{0.397} & 0.398          \\
\bottomrule
\end{tabular}

\label{tab:ablation_all_sets_patch_size}
\end{table*}

\begin{table*}[t]

\caption{All Datasets Merge Cat vs Merge Sum Ablation Study. The best results per row are marked in bold.}

\centering 
\footnotesize
\setlength{\tabcolsep}{1.5pt}

\begin{tabular}{lcccccccccccccccccccccccc}
\toprule
    Dataset & \multicolumn{2}{c}{Chamfer (x100)~$\Downarrow$} & \multicolumn{2}{c}{IoU~$\Uparrow$} & \multicolumn{2}{c}{F1~$\Uparrow$} & \multicolumn{2}{c}{Normal Error~$\Downarrow$} \\
    \cmidrule(lr){2-3} \cmidrule(lr){4-5} \cmidrule(lr){6-7} \cmidrule(lr){8-9} 
\name Merge              & Cat            & Sum            & Cat            & Sum            & Cat            & Sum            & Cat            & Sum            \\
\midrule       
ABC var. noise           & 0.661          & \textbf{0.659} & 0.834          & \textbf{0.835} & 0.898          & \textbf{0.899} & 0.305          & \textbf{0.301} \\ 
\midrule       
ABC no noise             & \textbf{0.495} & 0.497          & 0.902          & \textbf{0.903} & \textbf{0.940} & \textbf{0.940} & 0.203          & \textbf{0.199} \\
Famous no noise          & 0.411          & \textbf{0.419} & \textbf{0.907} & 0.906          & \textbf{0.947} & 0.945          & \textbf{0.487} & 0.491          \\
Thingi10k no noise       & 0.326          & \textbf{0.319} & 0.933          & \textbf{0.936} & 0.956          & \textbf{0.958} & 0.198          & \textbf{0.190} \\
\textbf{Mean no noise}   & \textbf{0.411} & 0.412          & 0.914          & \textbf{0.915} & \textbf{0.948} & \textbf{0.948} & 0.296          & \textbf{0.294} \\ 
\midrule       
Famous med. noise        & 0.497          & \textbf{0.485} & 0.879          & \textbf{0.883} & 0.930          & \textbf{0.933} & 0.555          & \textbf{0.542} \\
Thingi10k med. noise     & \textbf{0.382} & 0.384          & 0.910          & \textbf{0.912} & 0.943          & \textbf{0.944} & 0.249          & \textbf{0.244} \\
\textbf{Mean med. noise} & 0.440          & \textbf{0.434} & 0.894          & \textbf{0.897} & 0.937          & \textbf{0.939} & 0.402          & \textbf{0.393} \\ 
\midrule       
ABC high noise           & \textbf{0.956} & 0.972          & \textbf{0.760} & 0.759          & \textbf{0.847} & \textbf{0.847} & \textbf{0.403} & 0.406          \\
Famous high noise        & 1.026          & \textbf{0.992} & 0.732          & \textbf{0.749} & 0.835          & \textbf{0.845} & 0.756          & \textbf{0.701} \\
Thingi10k high noise     & \textbf{0.820} & 0.824          & 0.809          & \textbf{0.812} & 0.880          & \textbf{0.883} & 0.433          & \textbf{0.423} \\
\textbf{Mean high noise} & 0.934          & \textbf{0.929} & 0.767          & \textbf{0.773} & 0.854          & \textbf{0.858} & 0.531          & \textbf{0.510} \\ 
\midrule       
Famous sparse            & 0.691          & \textbf{0.629} & 0.848          & \textbf{0.853} & 0.911          & \textbf{0.915} & 0.619          & \textbf{0.600} \\
Thingi10k sparse         & 0.671          & \textbf{0.661} & 0.840          & \textbf{0.849} & 0.897          & \textbf{0.904} & 0.401          & \textbf{0.383} \\
\textbf{Mean sparse}     & 0.681          & \textbf{0.645} & 0.844          & \textbf{0.851} & 0.904          & \textbf{0.909} & 0.510          & \textbf{0.491} \\ 
\midrule       
Famous dense             & 0.422          & \textbf{0.404} & 0.907          & \textbf{0.908} & \textbf{0.949} & 0.947          & 0.492          & \textbf{0.488} \\
Thingi10k dense          & \textbf{0.322} & 0.326          & 0.932          & \textbf{0.933} & \textbf{0.959} & \textbf{0.959} & \textbf{0.209} & 0.212          \\
\textbf{Mean dense}      & 0.372          & \textbf{0.365} & 0.920          & \textbf{0.921} & \textbf{0.954} & 0.953          & \textbf{0.350} & \textbf{0.350} \\ 
\midrule       
\textbf{Mean overall}    & 0.591          & \textbf{0.582} & 0.861          & \textbf{0.864} & 0.915          & \textbf{0.917} & 0.408          & \textbf{0.398} \\
\bottomrule
\end{tabular}

\label{tab:ablation_all_sets_cat_sum}
\end{table*}